\documentclass{article}
\usepackage{spconf,amsmath,graphicx}
\usepackage{array}
\usepackage{wasysym}
\usepackage{url}
\usepackage{multirow,booktabs}
\usepackage{float}
\usepackage[caption = false]{subfig}
\title{MOTION BLUR REMOVAL VIA COUPLED AUTOENCODER}
\name{Kavya Gupta$^{\star}$ \qquad Brojeshwar Bhowmick$^{\star}$ \qquad Angshul Majumdar$^{\dagger}$}

\address{$^{\star}$ Embedded Systems and Robotics, TCS Research and Innovation, India \\
         $^{\dagger}$ Indraprastha Institute of Information Technology Delhi, India}

\begin{document}
%
\maketitle
\begin{abstract}
In this paper a joint optimization technique has been proposed for coupled autoencoder which learns the autoencoder weights and coupling map (between source and target) simultaneously. The technique is applicable to any transfer learning problem. In this work, we propose a new formulation that recasts deblurring as a transfer learning problem; it is solved using the proposed coupled autoencoder. The proposed technique can operate on-the-fly; since it does not require solving any costly inverse problem. Experiments have been carried out on state-of-the-art techniques; our method yields better quality images in shorter operating times.   
\end{abstract}
\section{Introduction}
\label{sec:intro}
Motion blurring is one of the common source of image corruption. It can be caused due to multiple reasons such as lens aberrations, camera shaking, long exposure time etc. The process of recovering sharp images from blurred images is termed as deblurring. This work proposes deblurring as a transfer learning problem, where the source domain is the blurred image and the target domain is the clean image. We solve it via coupled autoencoder. Our technique is generic and applicable for solving any inverse problem in imaging, e.g. denoising, inpainting, super-resolution etc. However, in this work we focus on deblurring. Despite significant progress in image deblurring, which is a well studied topic in computer vision, the existing deblurring algorithms often fail due to non-generalized kernel based approaches and also due to computational complexity. In this paper we recover the sharp image for a given blurred image with the help of a learning framework and hence can be generalized to broad class of blur. We focus on removal of the motion blur - uniform and non-uniform (space variant).

There are plethora of studies such as \cite{ren2015vectorization},\cite{xu2014deep},\cite{dong2016image},\cite{sun2015learning} which use neural networks and CNNs frameworks for solving computer vision tasks. But there are only a few studies \cite{cho2013simple},\cite{mao2016image},\cite{zeng2017coupled} which  solve inverse problems via autoencoders. In \cite{cho2013simple}, the authors used Sparse Stacked Autoencoders (SSAE) for denoising the images which can also be put in deblurring framework.
The motivation for having fast deblurring techniques comes from the requirement of sharp and undistorted images for the purpose of SLAM (Simultaneous Localization and Mapping), visual odometry , optical flow etc. There is a constant problem of motion blur while acquiring images and videos with the cameras fitted to the high speed motion drones. Distorted images will intervene with the mapping of the visual points, hence the pose estimation and tracking will be corrupted. Usual deblurring techniques solve an iterative inverse problem. This yields good quality images, but precludes itself from real-time applications. Our proposed method is light weighted, learning source and target autoencoders with mapping between source and target simultaneously. 

\section{Literature Review}
\label{review}
\subsection{Deblurring Techniqes}
Image deblurring techniques can be categorized into two types - blind and non-blind techniques. Non-blind techniques require priors about the blur kernel and it's parameters whereas for blind deblurring techniques we assume that the kernel is unknown. Estimation of accurate kernels is detrimental for good deblurring especially in case of space variant blurs. Single image deblurring techniques jointly estimate the motion kernels and sharp image. \cite{krishnan2011blind},\cite{pan2014deblurring},\cite{xu2013unnatural} use sparsity priors to retreive latent sharp image for better kernel estimations. In \cite{whyte2014deblurring}, blur kernel is estimated in the camera motion space itself.

Recent development has been  on devising learning  based techniques for learning the degradation models. In \cite{xu2014deep} authors proposed an image deconvolution neural network for non-blind deconvolution which focuses on removal of uniform blur. In \cite{ren2015vectorization} a vectorization based CNN method was proposed which showed improvement on various high and low level vision tasks. \cite{sun2015learning} proposes a deep learning approach of predicting the motion kernel at patch level using a CNN. 
\subsection{Coupled Representation Learning}
Coupled dictionary learning has a rich literature \cite{wang2012semi},\cite{yang2012coupled},\cite{huang2013coupled}. It has been applied to a wide range of problems in image synthesis, e.g., single image super-resolution, photo-sketch synthesis, cross spectral (RGB-NIR) face recognition, RGB-depth classification etc. It has also been used for trans-lingual information retrieval \cite{mehrotratowards}. The main idea in coupled dictionary is learning of dictionaries (and corresponding coefficients) for the two domains - source and target, such that the coefficients from one domain can be linearly mapped to the other.
	
The concept of coupled autoencoder is new; it follows from dictionary learning. The main idea here is to learn an autoencoder for the source and another for the target along with a mapping from the source to the target (semi-coupled) and vice versa (fully coupled). There are only a handful of studies on this. In \cite{zeng2017coupled}, it learns two deep stacked autoencoders for two domains - source (low resolution image) and target (high resolution image). These are learnt separately. Once the autoencoders are learnt a mapping from the deepest layer of the source autoencoder is learnt to the target autoencoder. The approach is piecemeal and hence sub-optimal. The autoencoders are learnt independently for the different domains and hence there is no feedback from one to the other during the learning stage. The mapping from the source to target is a stand-alone process which has no influence on the source and target autoencoder learning. 

In another study \cite{wang2014deeply}, it learns coupled shallow autoencoders and stacks them up to form a deep architecture greedily. As in the prior work, their coupled autoencoder learning process is sub-optimal as well. The autoencoders (for source and target) are learnt separately; finally a mapping between the two is learnt, consequently the coupling has no bearing on the autoencoder training.

The only prior work that optimally learns the mapping during the autoencoder training process is \cite{wang2016coupled}. However they use a marginalized denoising autoencoder, which is much simpler to solve compared to the full autoencoder having separate encoding and decoding layers. 

\section{Proposed Formulation}
\label{sec:Proposed}
This is the first work that introduces an optimal formulation for coupled autoencoder. Unlike prior studies, the autoencoders for source and target will be learnt along with the linear mapping between the two. This is optimal in the sense that all the variables influence each other in the learning process - this has been missing in prior works. 
\begin{figure}[h!]	
	\includegraphics[width = 8.5 cm,height= 6 cm]{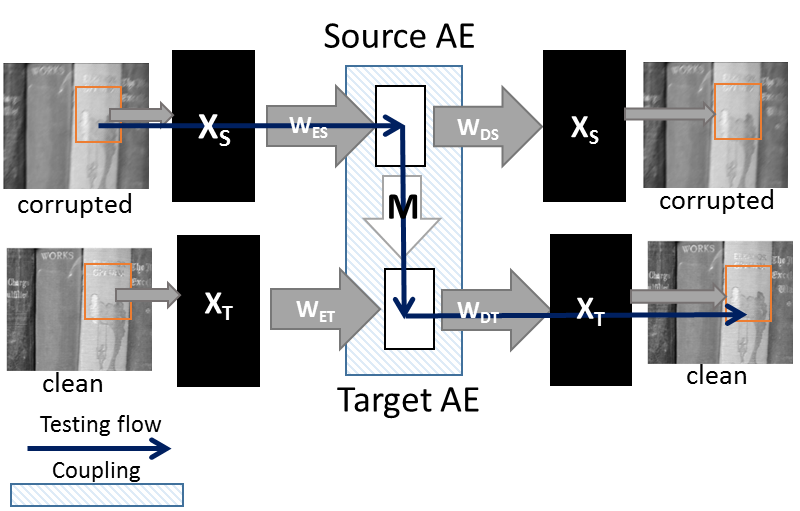}
	\centering
	\caption{Schematic Diagram of Coupled Autoencoder}
	\label{fig:CAE}
\end{figure}
Figure \ref{fig:CAE} shows the schematic diagram. The source autoencoder uses the blurred samples and the target autoencoder uses the corresponding clean samples. The coupling learns to map the representation from the source (blurred) to the target (clean). Mathematically this is expressed as,
\begin{equation}\label{1}
\begin{split}
\underset{W_{DS},W_{ES},W_{DT},W_{ET},M}{\text{argmin}}  {\parallel X_S-W_{DS}\varphi(W_{ES}X_S) \parallel}_{F}^{2} \\+ {\parallel X_T-W_{DT}\varphi(W_{ET}X_T) \parallel}_{F}^{2}   \\+ \lambda{\parallel \varphi(W_{ET}X_T)-M\varphi(W_{ES}X_S) \parallel}_{F}^{2}
\end{split}
\end{equation}
The first term ${\parallel X_S-W_{DS}\varphi(W_{ES}X_S) \parallel}_{F}^{2}$ is the standard autoencoder formulation for the source. Here $W_{DS}$ corresponds to the decoder and $W_{ES}$ is the encoder. The second term ${\parallel X_T-W_{DT}\varphi(W_{ET}X_T) \parallel}_{F}^{2}$ is the autoencoder for the target; $W_{DT}$ and $W_{ET}$ are the decoder and the encoder respectively. ${\parallel \varphi(W_{ET}X_T)-M\varphi(W_{ES}X_S) \parallel}_{F}^{2}$is the coupling term. The linear map M couples the representation from the source to that of the target. 

We solve it using the variable splitting technique. The first step is to introduce proxies for the representations, i.e. $Z_S = \varphi(W_{ES}X_S)$ and $Z_T = \varphi(W_{ET}X_T)$ and forming the Lagrangian.
A better approach to handle this is to use the Split Bregman technique \cite{goldstein2009split}. A Bregman relaxation variable is introduced such that the value of the hyper-parameter need not be changed; the relaxation variable is updated in every iteration automatically so that it enforces equality at convergence. The Split Bregman formulation is,
\begin{table*}[htbp]
	\centering
	\resizebox{\textwidth}{!}{
	\begin{tabular}{|cc|c|c|c|c|c|c|c|c|}
		\hline
		Blur type & &Krishnan et al.\cite{krishnan2011blind}&Whyte et al.\cite{whyte2014deblurring}&Pan et al.\cite{pan2014deblurring}&Xu et al.\cite{xu2013unnatural}&Xu et al.\cite{xu2010two} & Ren et al.\cite{ren2015vectorization}&Proposed \\
		\hline
		\multirow{2}{*}{\small{Uniform blur}}&\tiny{(PSNR)}&23.7679&23.5257&23.6927&22.9265&25.6540&20.9342&\bf{30.8893} \\&\tiny{(SSIM)}&0.6929&0.6899&0.7015&0.6620&0.7708&0.7312&\bf{0.8787} \\
		\hline
		\multirow{2}{*}{\small{Non Uniform blur}}&\tiny{(PSNR)}&20.3013&20.4161&19.6594&20.5548&19.9718&20.8226&\bf{29.6364} \\&\tiny{(SSIM)}&0.5402&0.5361&0.5345&0.5812&0.5692&0.7328&\bf{0.8711} \\
		\hline
	\end{tabular}
	}
	\caption{Comparison of PSNR and SSIM values}
	\label{table}
\end{table*}
\begin{equation}\label{4}
\begin{split}
\underset{W_{DS},W_{ES},W_{DT},W_{ET},M,Z_S,Z_T}{\text{argmin}}{\parallel X_S-W_{DS}Z_S \parallel}_{F}^{2} \\+ {\parallel X_T-W_{DT}Z_T \parallel}_{F}^{2} \\+ \lambda{\parallel Z_T-MZ_S \parallel}_{F}^{2} \\+ \eta{{\parallel Z_S-\varphi(W_{ES}X_S)-B_S \parallel}_{F}^{2}}\\+\eta{{\parallel Z_T-\varphi(W_{ET}X_T)-B_T \parallel}_{F}^{2}}
\end{split} 
\end{equation}
Here $B_S$ and $B_T$ are Bregman relaxation variables. 
We invoke the alternating minimization method of multipliers \cite{boyd2011distributed} to segregate (\ref{4}) into the following (simpler) sub-problems.\\\\
P1 : $\underset{W_{DS}}{\text{argmin}} {\parallel X_S-W_{DS}Z_S \parallel}_{F}^{2}$\\
P2 : $\underset{W_{DT}}{\text{argmin}} {\parallel X_T-W_{DT}Z_T \parallel}_{F}^{2}$ \\
P3 : $\underset{W_{ES}}{\text{argmin}} {\parallel Z_S-\varphi(W_{ES}X_S)-B_S \parallel}_{F}^{2} \\\hspace*{1.5 em}\equiv \underset{W_{ES}}{\text{argmin}} {\parallel \varphi^{-1}(Z_S-B_S)-W_{ES}X_S \parallel}_{F}^{2}$ \\
P4 : $\underset{W_{ET}}{\text{argmin}} {\parallel Z_T-\varphi(W_{ET}X_T)-B_T \parallel}_{F}^{2} \\\hspace*{1.5 em}\equiv  \underset{W_{ET}}{\text{argmin}} {\parallel \varphi^{-1}(Z_T-B_T)-W_{ET}X_T \parallel}_{F}^{2}$ \\
P5 : $\underset{Z_{S}} {\text{argmin}} {\parallel X_S-W_{DS}Z_S \parallel}_{F}^{2}+\lambda{\parallel Z_T-MZ_S \parallel}_{F}^{2}+ \\\hspace*{2.5 em}\eta{\parallel Z_S-\varphi(W_{ES}X_S)-B_S \parallel}_{F}^{2} $ \\
P6 : $\underset{Z_{T}} {\text{argmin}} {\parallel X_T-W_{DT}Z_T \parallel}_{F}^{2}+\lambda{\parallel Z_T-MZ_S \parallel}_{F}^{2}+ \\\hspace*{2.5 em}\eta{\parallel Z_T-\varphi(W_{ET}X_T)-B_T \parallel}_{F}^{2} $ \\
P7 : $\underset{M}{\text{argmin}}{\parallel Z_T-MZ_S \parallel}_{F}^{2}$\\

Sub-problems P1, P2, P5, P6, P7 are simple least squares problems having an analytic solution in the form of pseudo-inverse. Sub-problems P3 and P4 are originally non-linear least squares problems; but since the activation function is trivial to invert (it is applied element-wise), we can convert P3 and P4 to simple linear least squares problems and solve them using pseudo-inverse.
The final step in every iteration is to update the Bregman relaxation variables. This is achieved by simple gradient descent.
\begin{equation}
B_S \leftarrow \varphi(W_{ES}X_S)+B_S-Z_S
\end{equation}
\begin{equation}
B_T \leftarrow \varphi(W_{ET}X_T)+B_T-Z_T
\end{equation}

We have used two stopping criteria. Iterations stop when a specified maximum number of iterations is reached or when the objective function converges to a local minimum. By convergence we mean that the value of the objective function does not change much in successive iterations. 

\section{Results}
\label{Results}

For the experimental evaluation of our proposed method, we use images from standard image blur dataset \cite{Mavridaki2014NoreferenceBA}- CERTH dataset. It contains 630 undistorted, 220 naturally-blurred and 150 artificially-blurred images in the training set and 619 undistorted, 411 naturally blurred and 450 artificially blurred images in the evaluation set. For performance evaluation we take a small subset of the dataset. We take 50 images for training, 20 images for testing and corrupt them with the motion blur kernels. The algorithm is tested for two scenarios – uniform kernel and  non-uniform kernel \cite{levin2009understanding}. The images in the dataset have huge sizes so we did the deblurring patch wise, a protocol followed in \cite{cho2013simple}. For training and testing we divided the whole image into number of overlapping patches. Each individual patch acts as a sample. We do not do any preprocessing to the images or patches. We use the blurry patches to train the source autoencoder, the corresponding clean patches to train the target autoencoder and a mapping is learnt between the target and source representation. During testing, the blurred patch is given as the input of the source. The corresponding source representation is obtained from the source encoder from which the target representation is obtained by the coupling map. Once the target representation is obtained, the decoder of the target is used to generate the clean (deblurred) patch. Once all the individual patches of the testing image are deblurred we reconstruct the whole image by placing patches at their locations and averaging at the overlapping regions. We do not get blocking artefacts consequently we do not do any post processing to the recovered images.

We use patches of 40x40 with an overlapping of 20x20 for all the images. The input size for the autoencoders hence become 1600 and are learned with 1400 hidden nodes. There are just two parameters $\lambda$ and $\eta$ while training which do not require much tuning. Whereas the testing is parameter free. The testing flow is shown in the Figure \ref{fig:CAE}.

We compare the proposed method with several state-of-the-art techniques with \cite{krishnan2011blind},\cite{pan2014deblurring},\cite{xu2013unnatural},\cite{whyte2014deblurring},\cite{xu2010two} which estimates the kernel and retreive deblurred image by deconvolution and \cite{ren2015vectorization}  which learns a Vectorized CNN for deblurring. The consolidated results are shown in Table \ref{table}. The PSNR and SSIM values are averaged over for all the 20 testing images. We see that we consistently perform better than all the methods on both PSNR and SSIM for both uniform and non-uniform blur. Our experimentation showed that the framework can learn characteristics for different kinds of motion blur and yields good performances on them as well. 

\begin{figure*}[h!]
	\centering
	\subfloat[]{\includegraphics[width = 1.35 in, height =1.5 in]{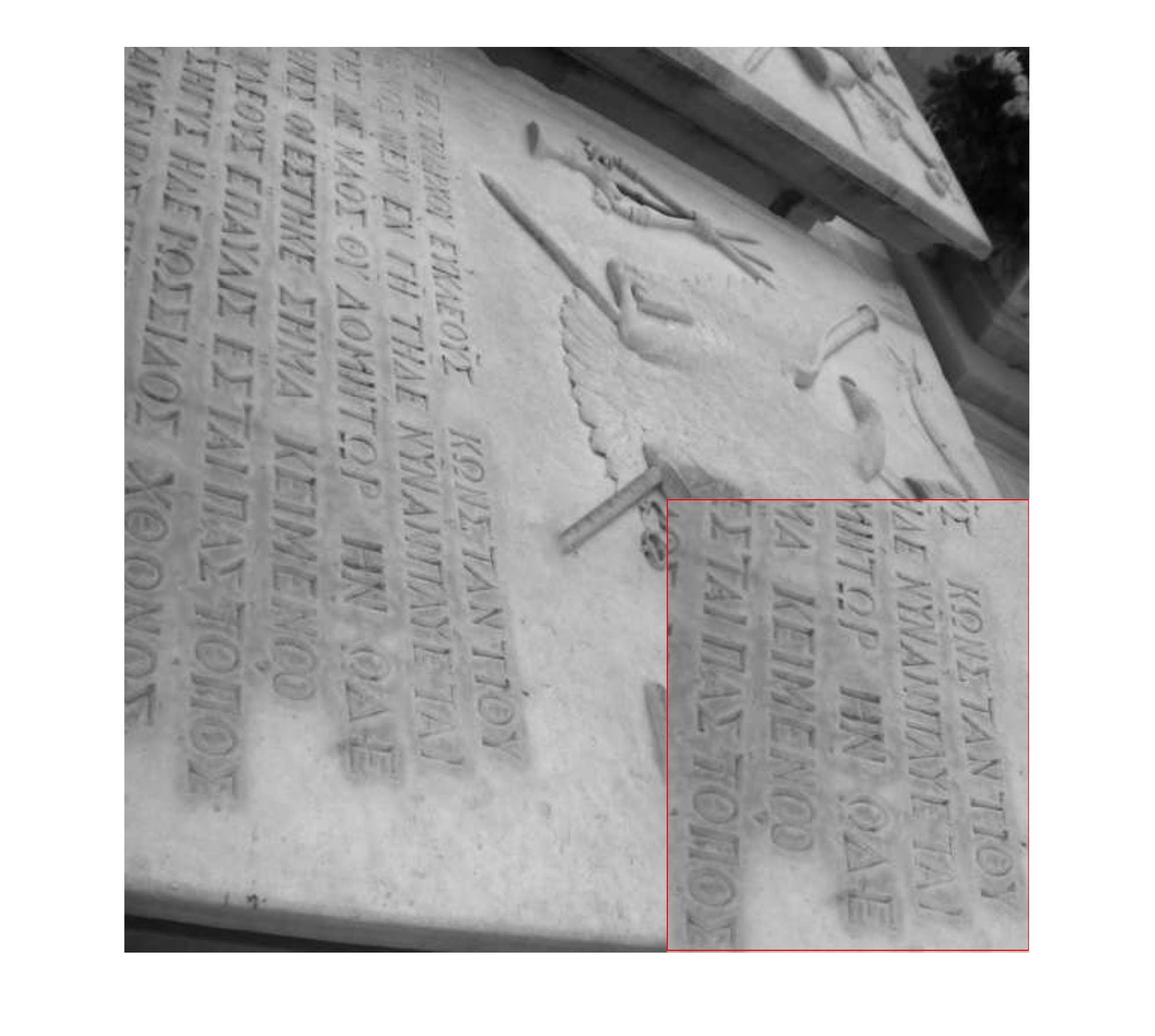}} 
	\subfloat[]{\includegraphics[width = 1.35 in, height =1.5 in]{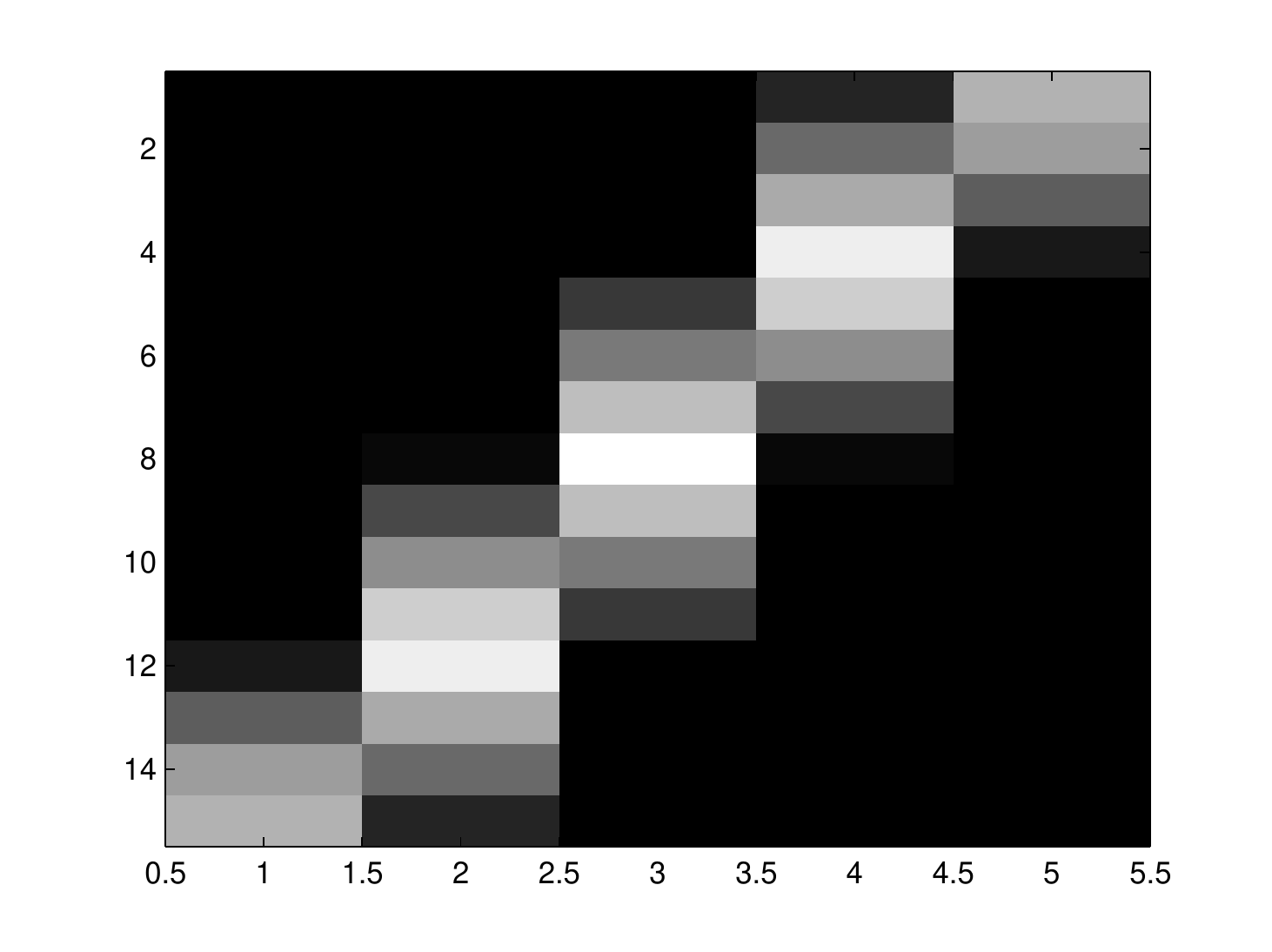}}
	\subfloat[]{\includegraphics[width = 1.35 in, height =1.5 in]{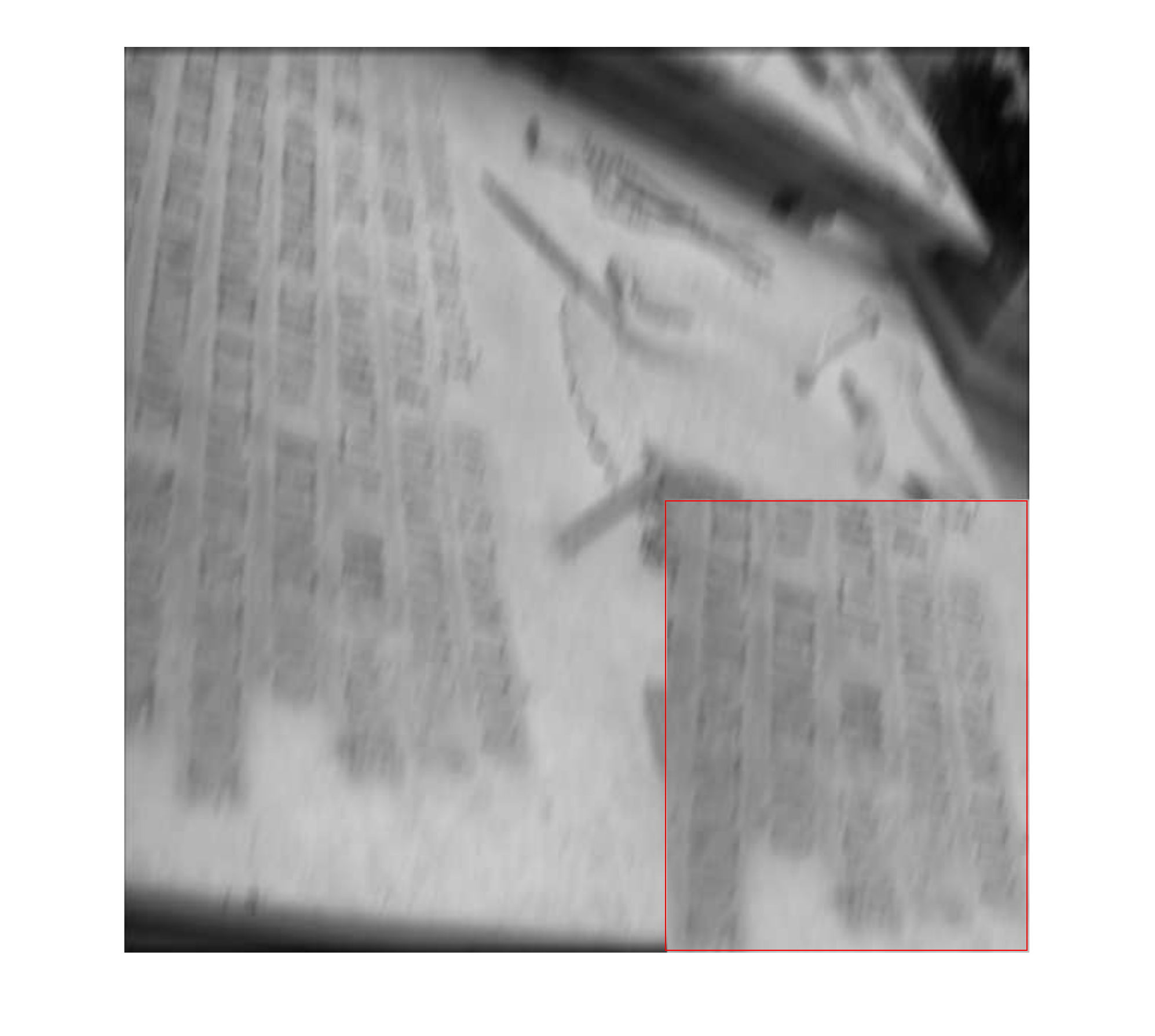}}
	\subfloat[\scriptsize{PSNR=25.80 SSIM=0.66}]{\includegraphics[width = 1.35 in, height =1.5 in]{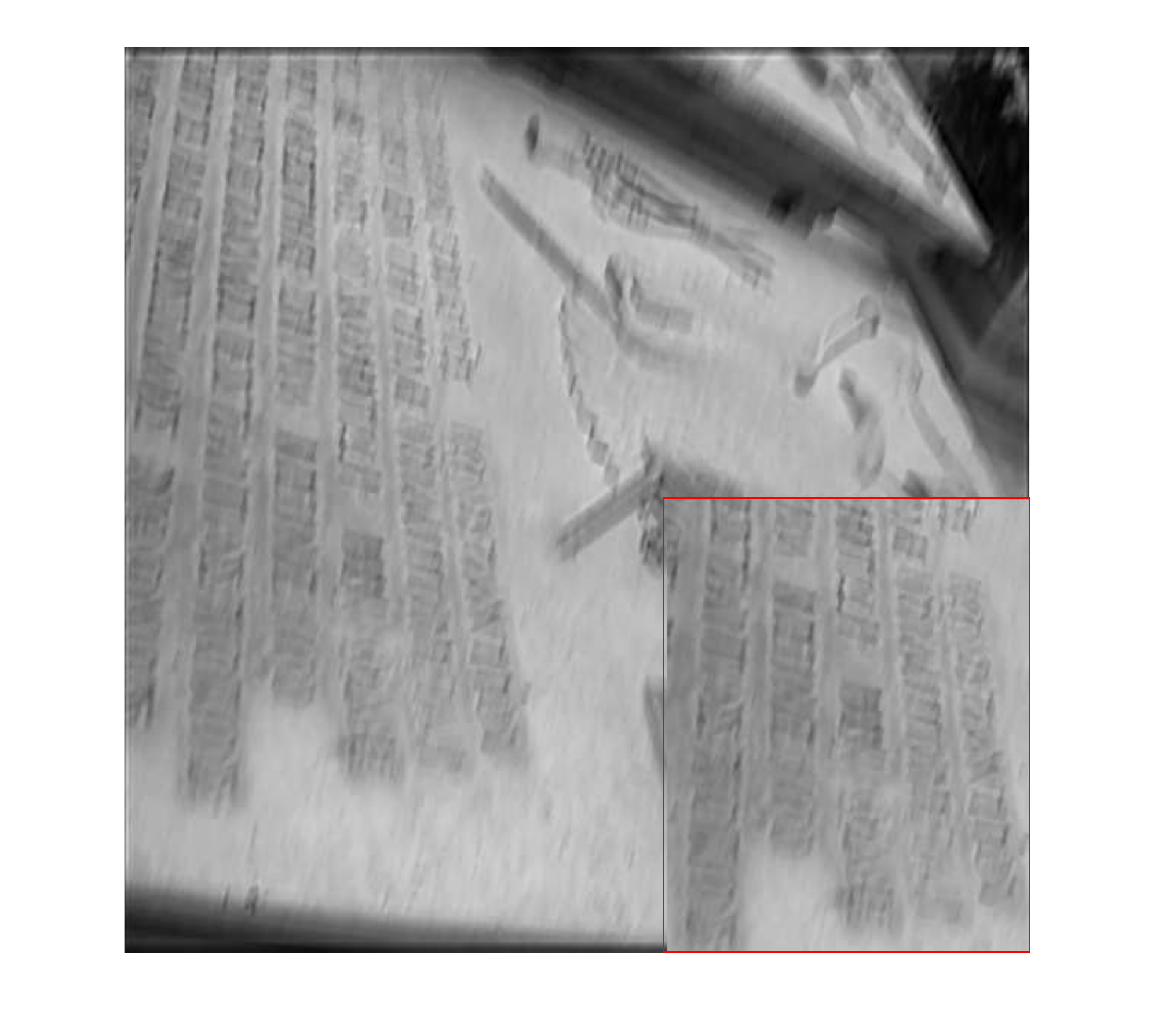}}
	\subfloat[\scriptsize{PSNR=25.19 SSIM=0.67}]{\includegraphics[width = 1.35 in, height =1.5 in]{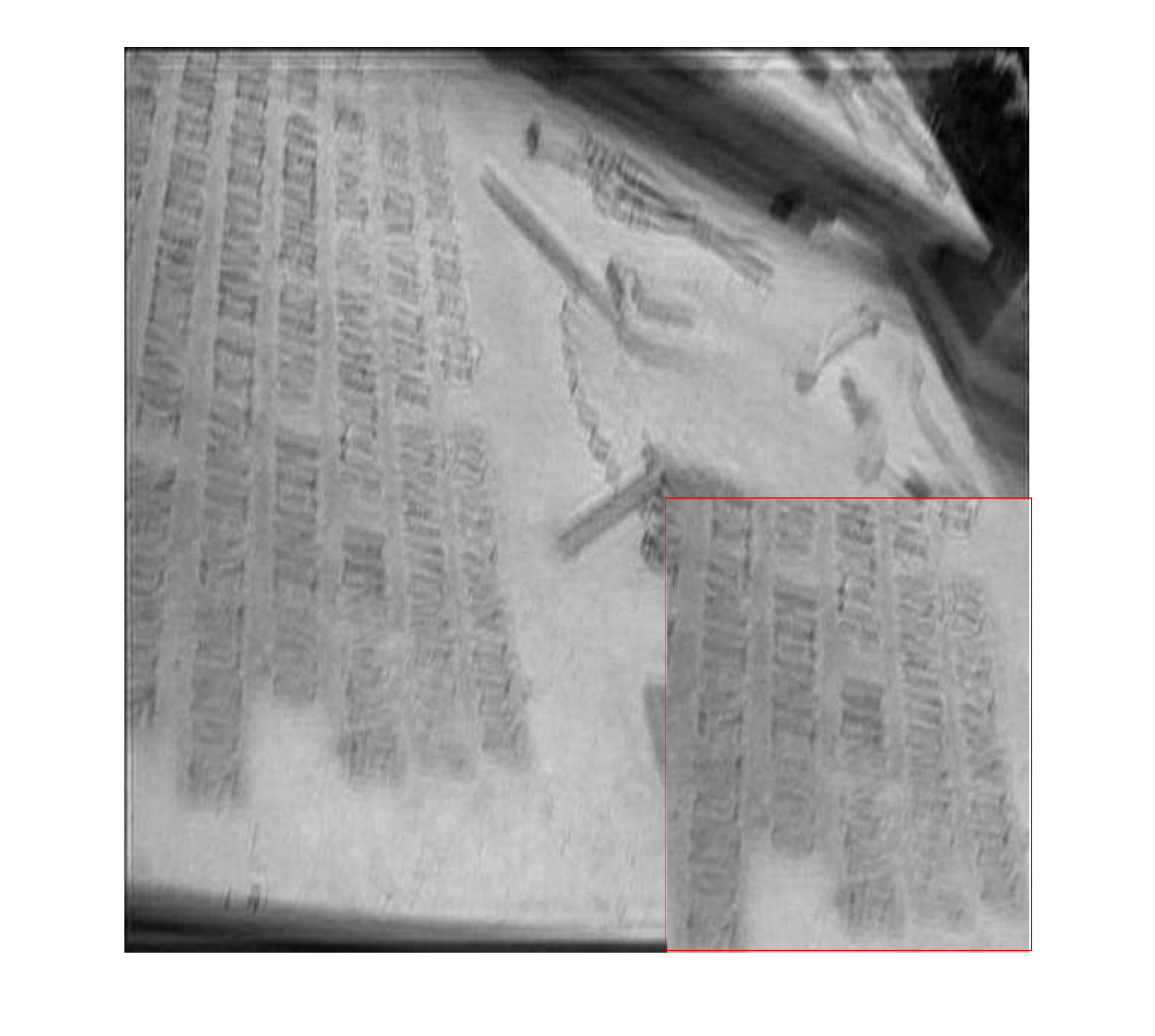}}\\
	\subfloat[\scriptsize{PSNR=26.13 SSIM=0.74}]{\includegraphics[width = 1.35 in, height =1.5 in]{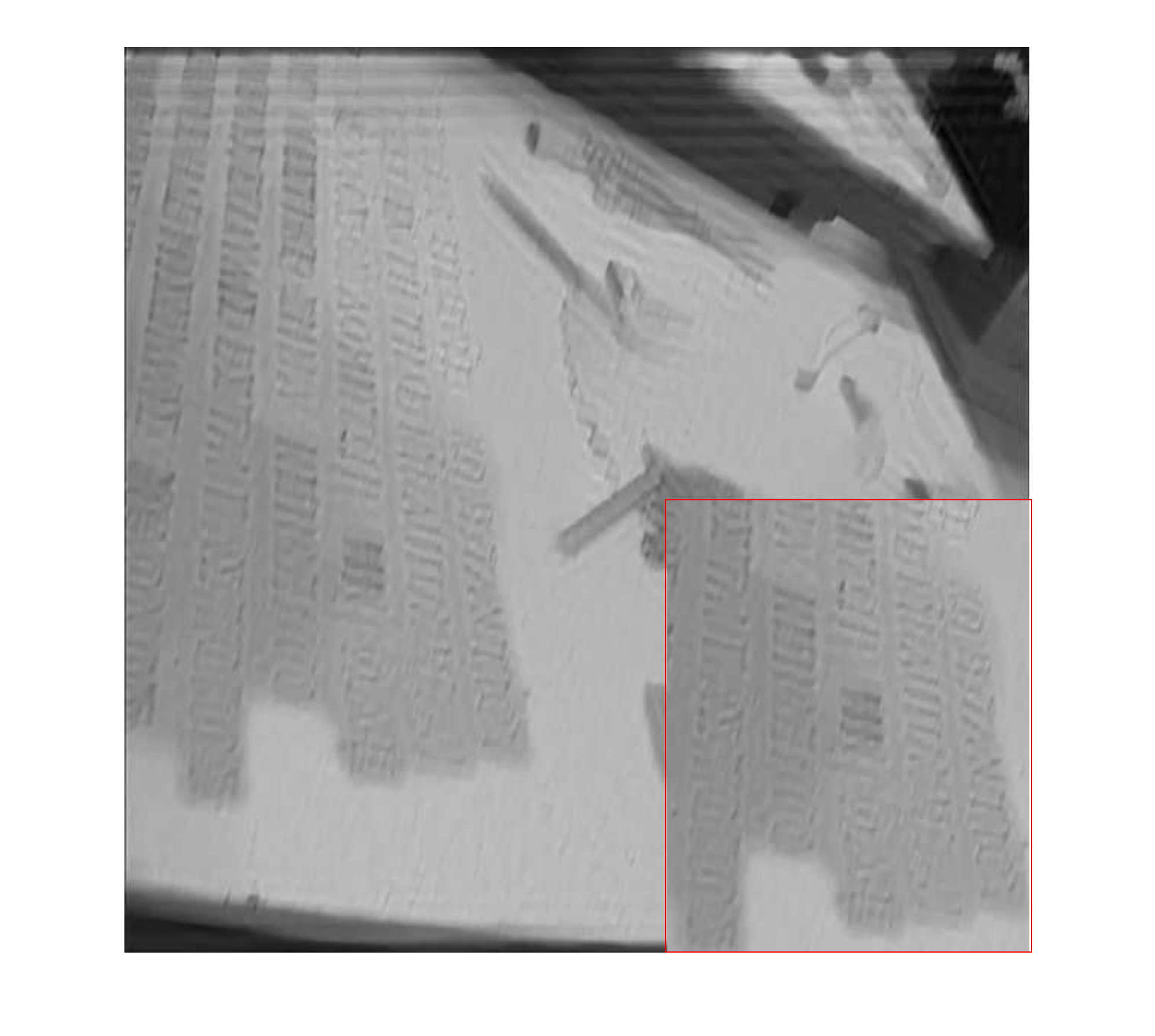}} 
	\subfloat[\scriptsize{PSNR=25.78 SSIM=0.66}]{\includegraphics[width = 1.35 in, height =1.5 in]{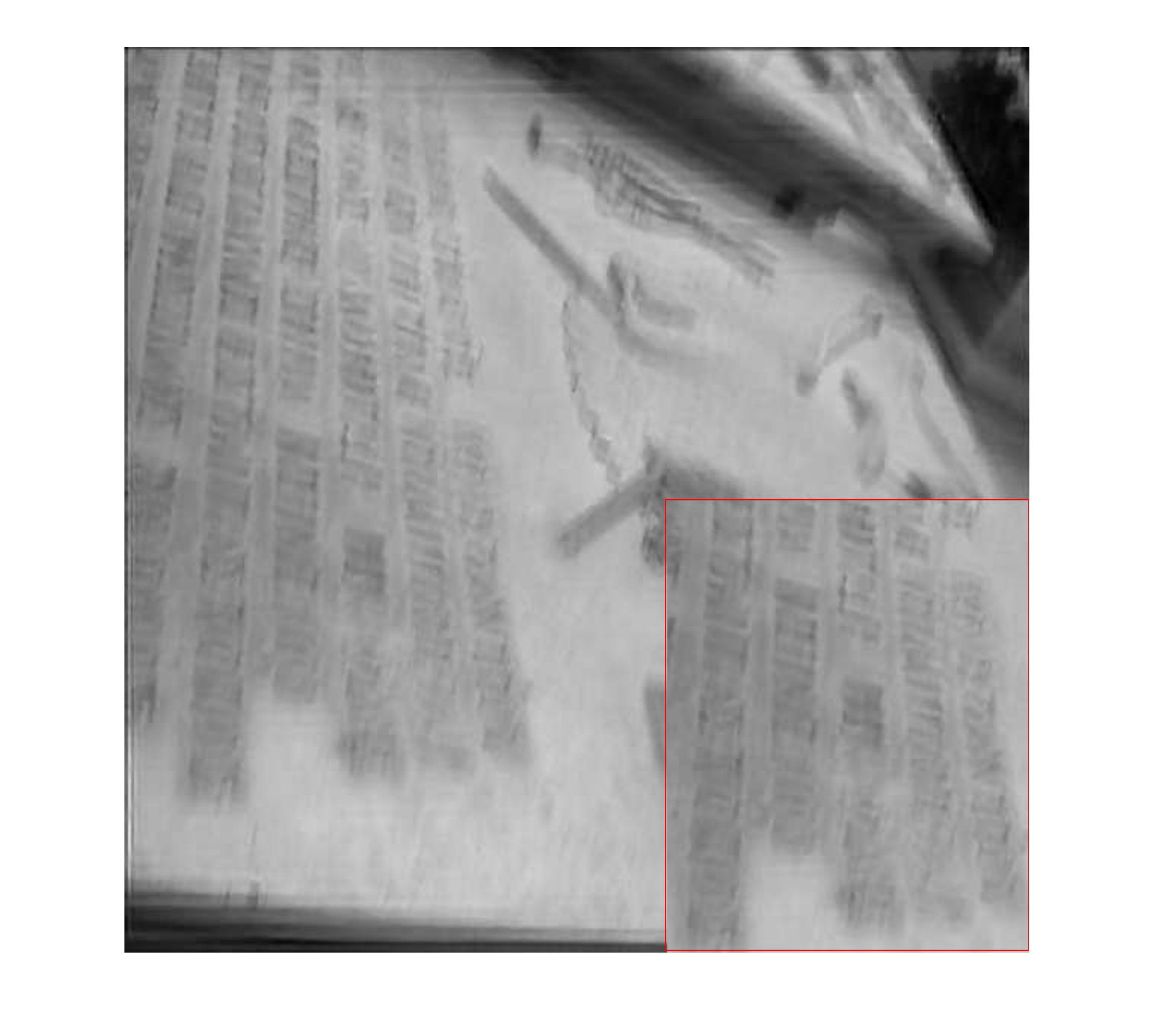}} 
	\subfloat[\scriptsize{PSNR=22.66 SSIM=0.65}]{\includegraphics[width = 1.35 in, height =1.5 in]{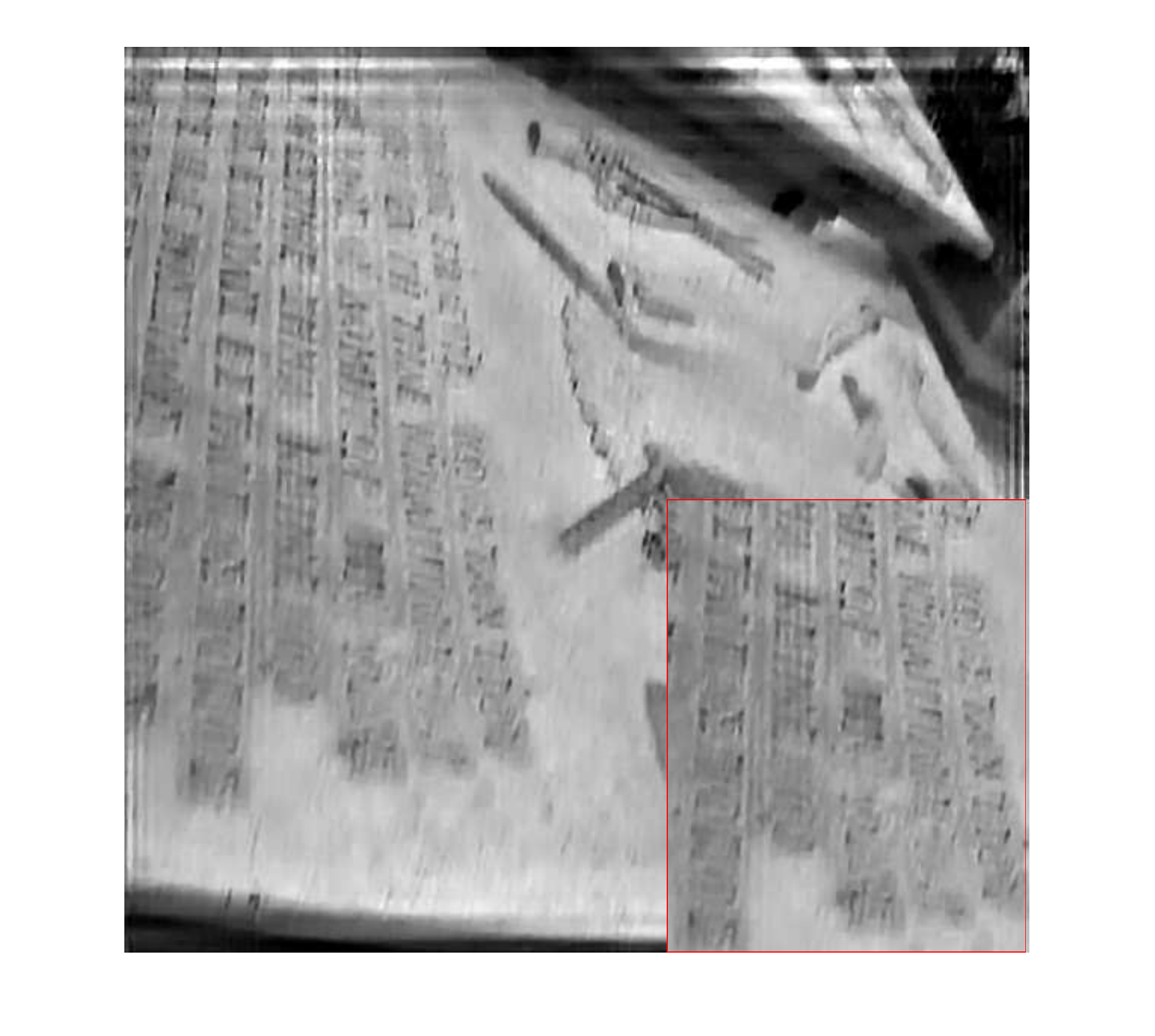}} 
	\subfloat[\scriptsize{PSNR=21.77 SSIM=0.74}]{\includegraphics[width = 1.35 in, height =1.5 in]{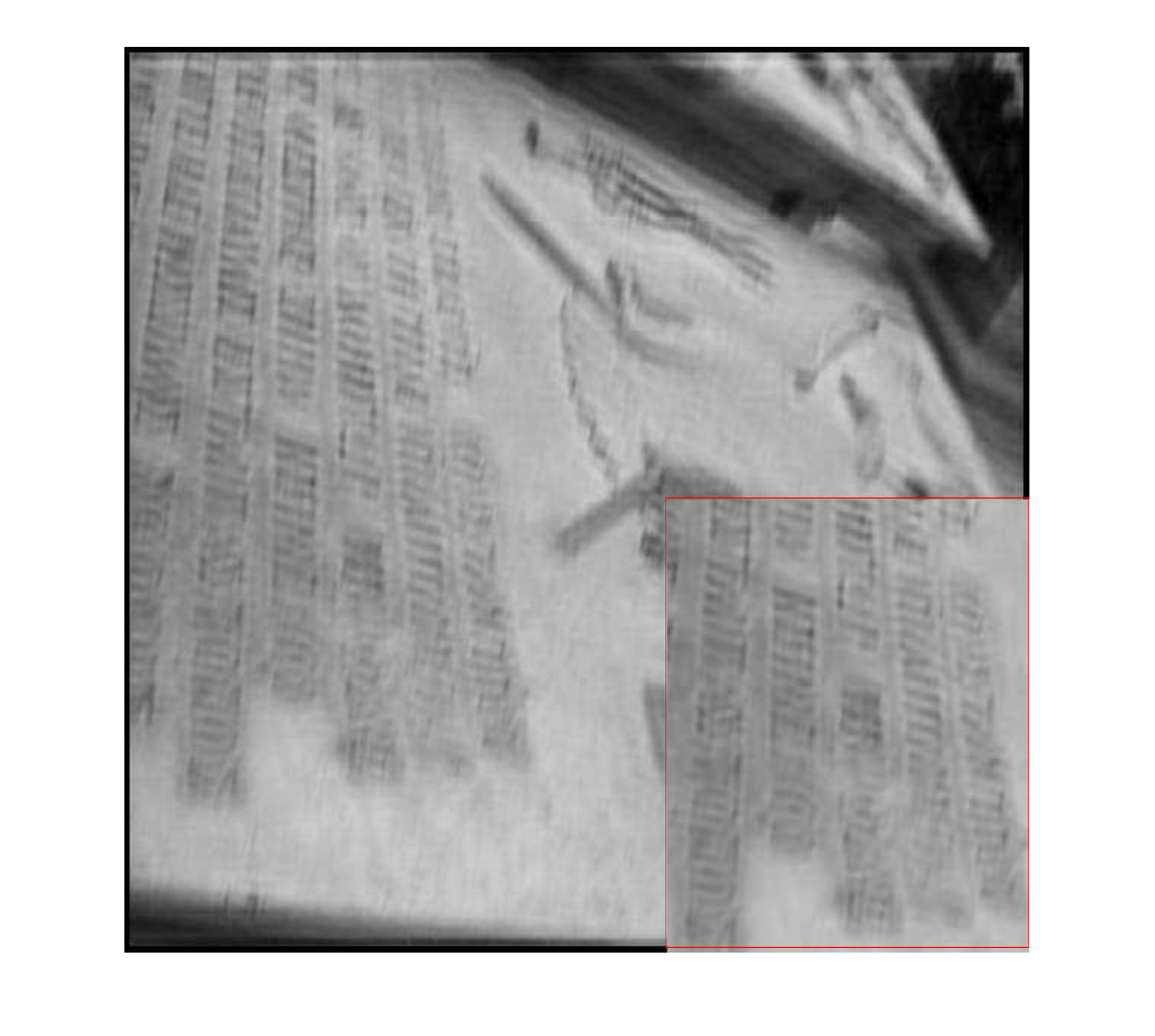}} 
	\subfloat[\scriptsize{PSNR=33.43 SSIM=0.90}]{\includegraphics[width = 1.35 in, height =1.5 in]{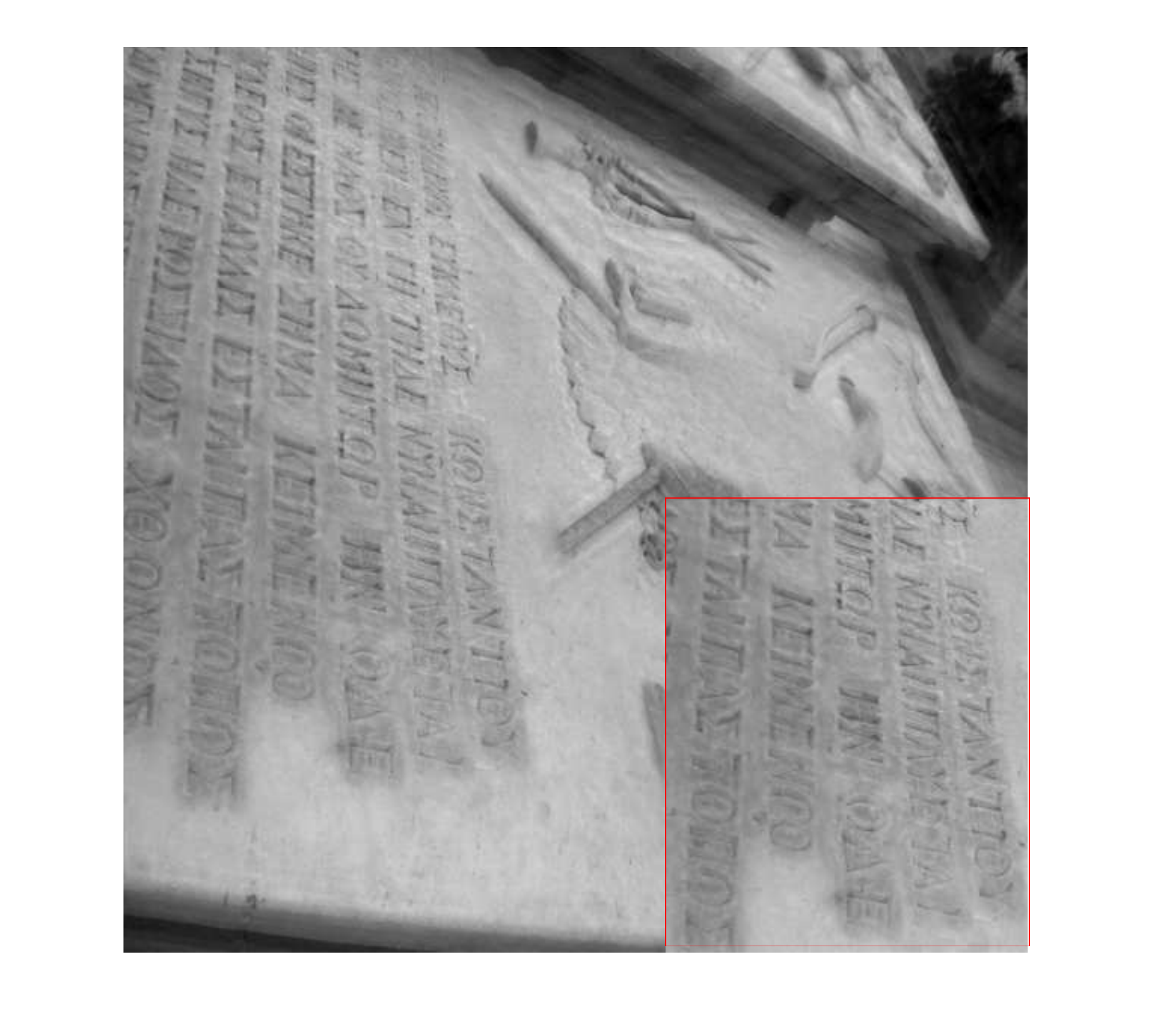}} 
	\caption{Uniform Blur Removal Comparison. (a) Ground truth (b) Blur kernel (c) Blurred (d) Krishnan et al. \cite{krishnan2011blind} (e) Whyte et al. \cite{whyte2014deblurring} (f) Pan et al. \cite{pan2014deblurring} (g) Xu et al. \cite{xu2013unnatural} (h) Xu et al. \cite{xu2010two} (i) Ren et al. \cite{ren2015vectorization} (j) Proposed }
	\label{Uniform}
\end{figure*}

\begin{figure*}[h!]
	\centering
	\subfloat[]{\includegraphics[width = 1.35 in, height =1.5 in]{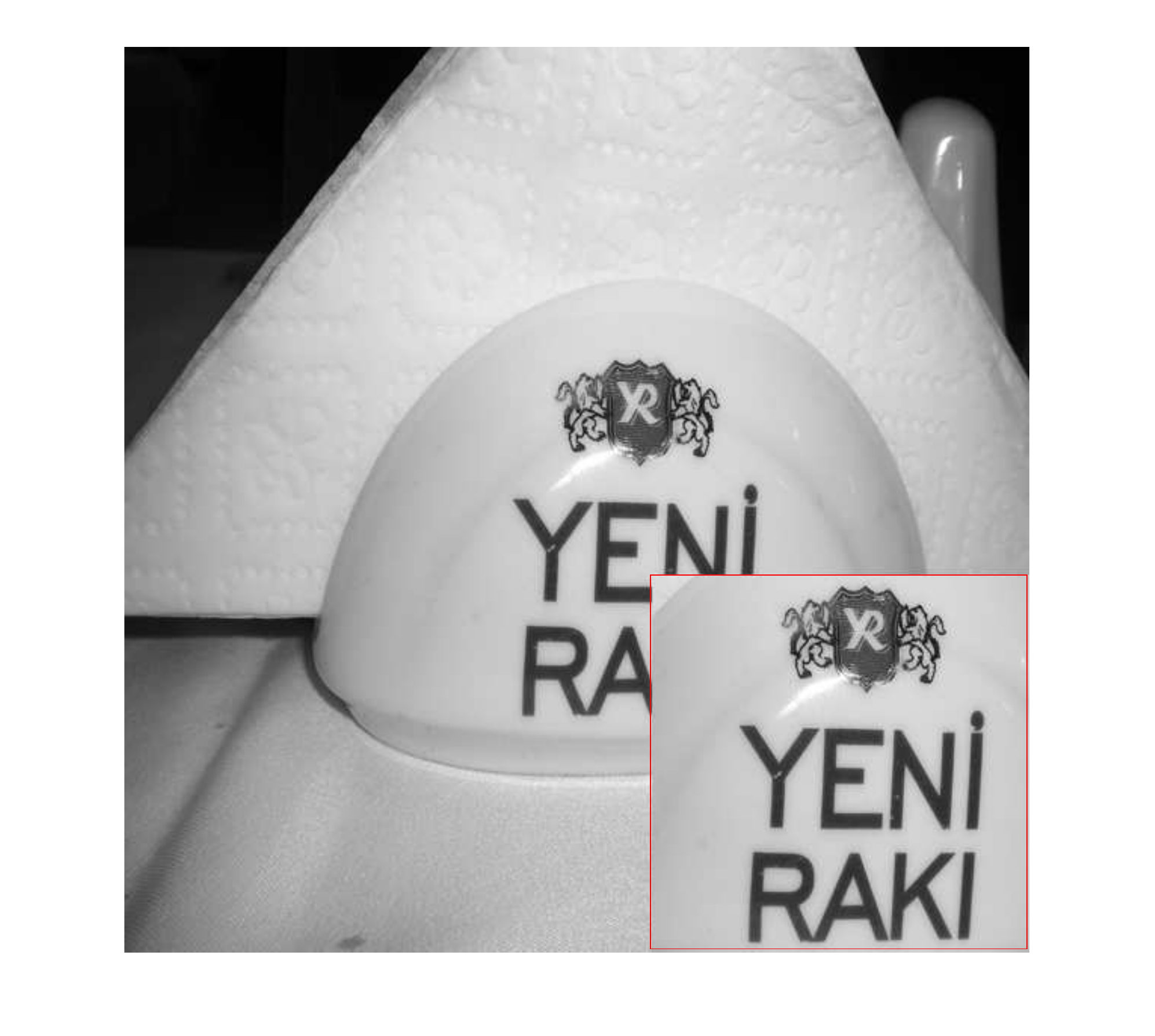}} 
	\subfloat[]{\includegraphics[width = 1.35 in, height =1.5 in]{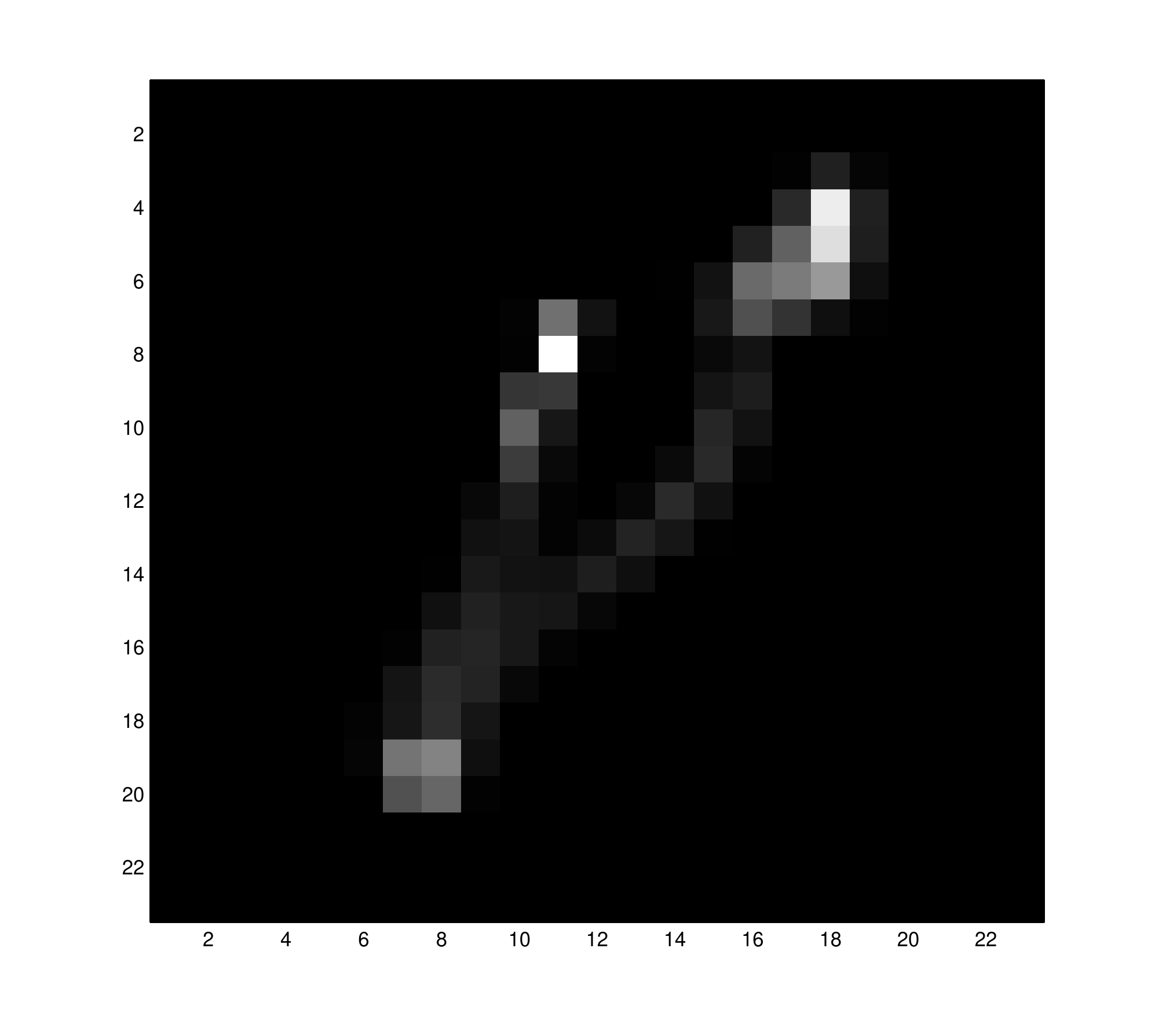}}
	\subfloat[]{\includegraphics[width = 1.35 in, height =1.5 in]{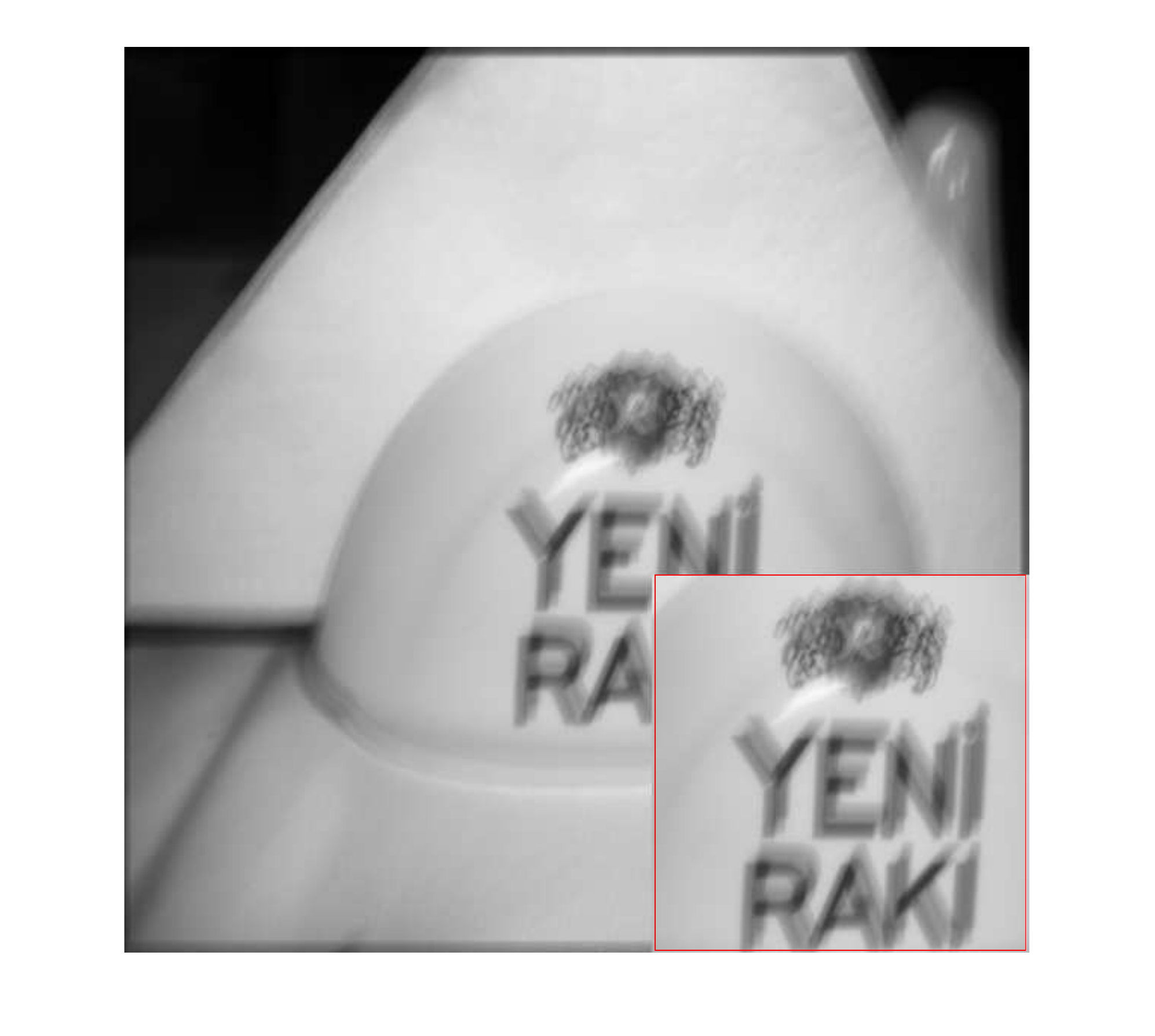}}
	\subfloat[\scriptsize{PSNR=21.73 SSIM=0.73}]{\includegraphics[width = 1.35 in, height =1.5 in]{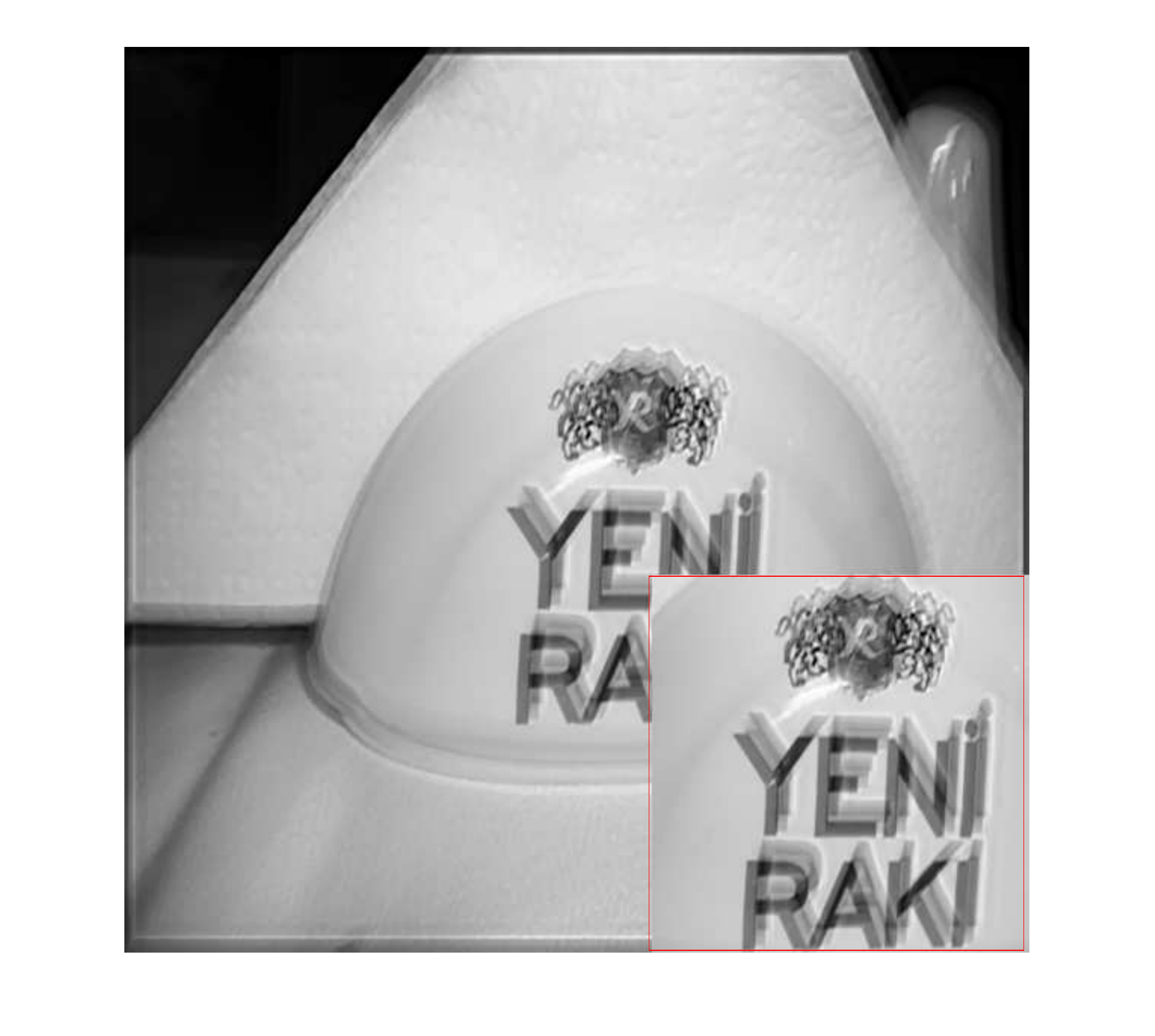}}
	\subfloat[\scriptsize{PSNR=21.47 SSIM=0.72}]{\includegraphics[width = 1.35 in, height =1.5 in]{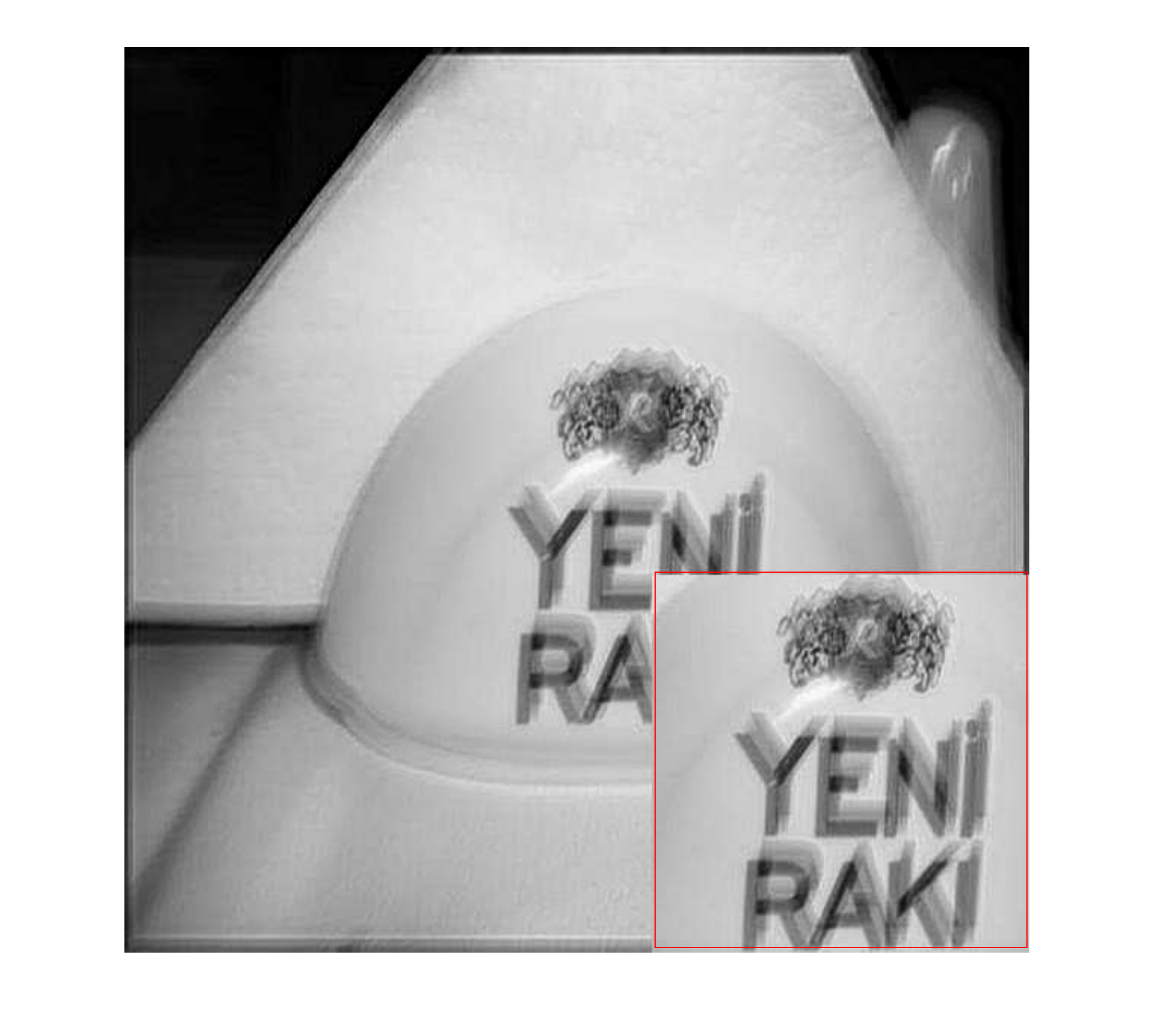}}\\
	\subfloat[\scriptsize{PSNR=21.43 SSIM=0.73}]{\includegraphics[width = 1.35 in, height =1.5 in]{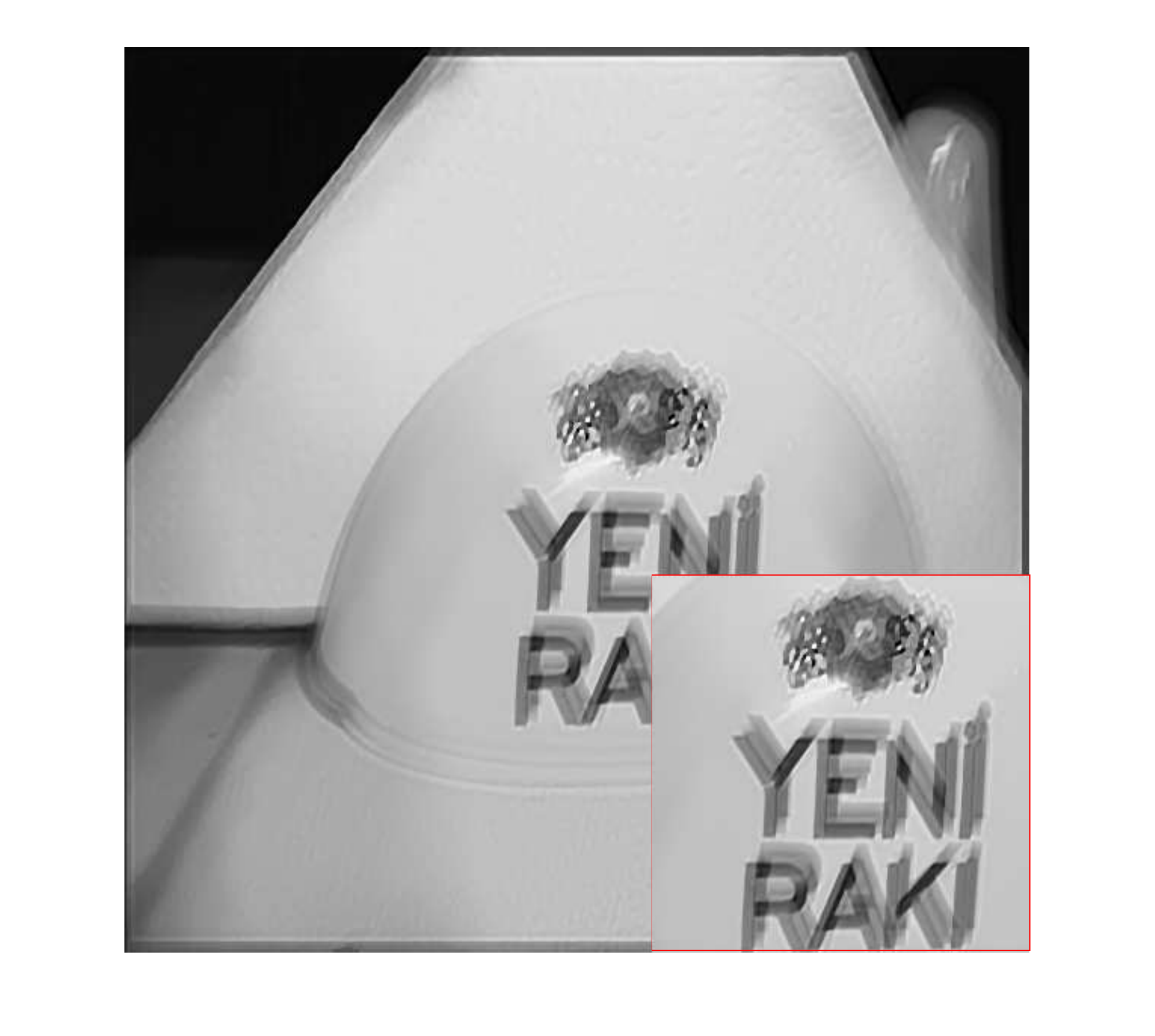}} 
	\subfloat[\scriptsize{PSNR=21.91 SSIM=0.76}]{\includegraphics[width = 1.35 in, height =1.5 in]{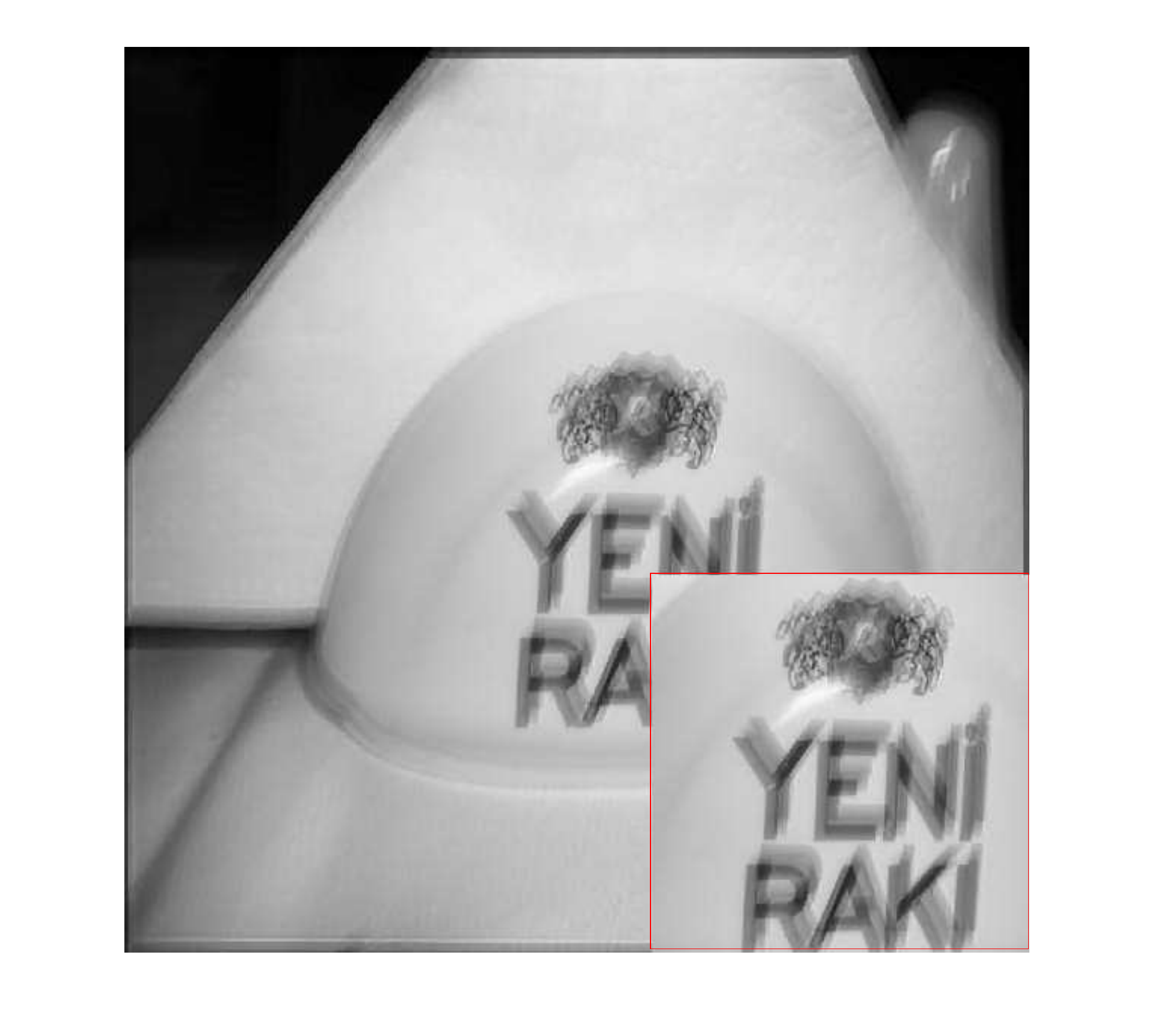}} 
	\subfloat[\scriptsize{PSNR=12.55 SSIM=0.45}]{\includegraphics[width = 1.35 in, height =1.5 in]{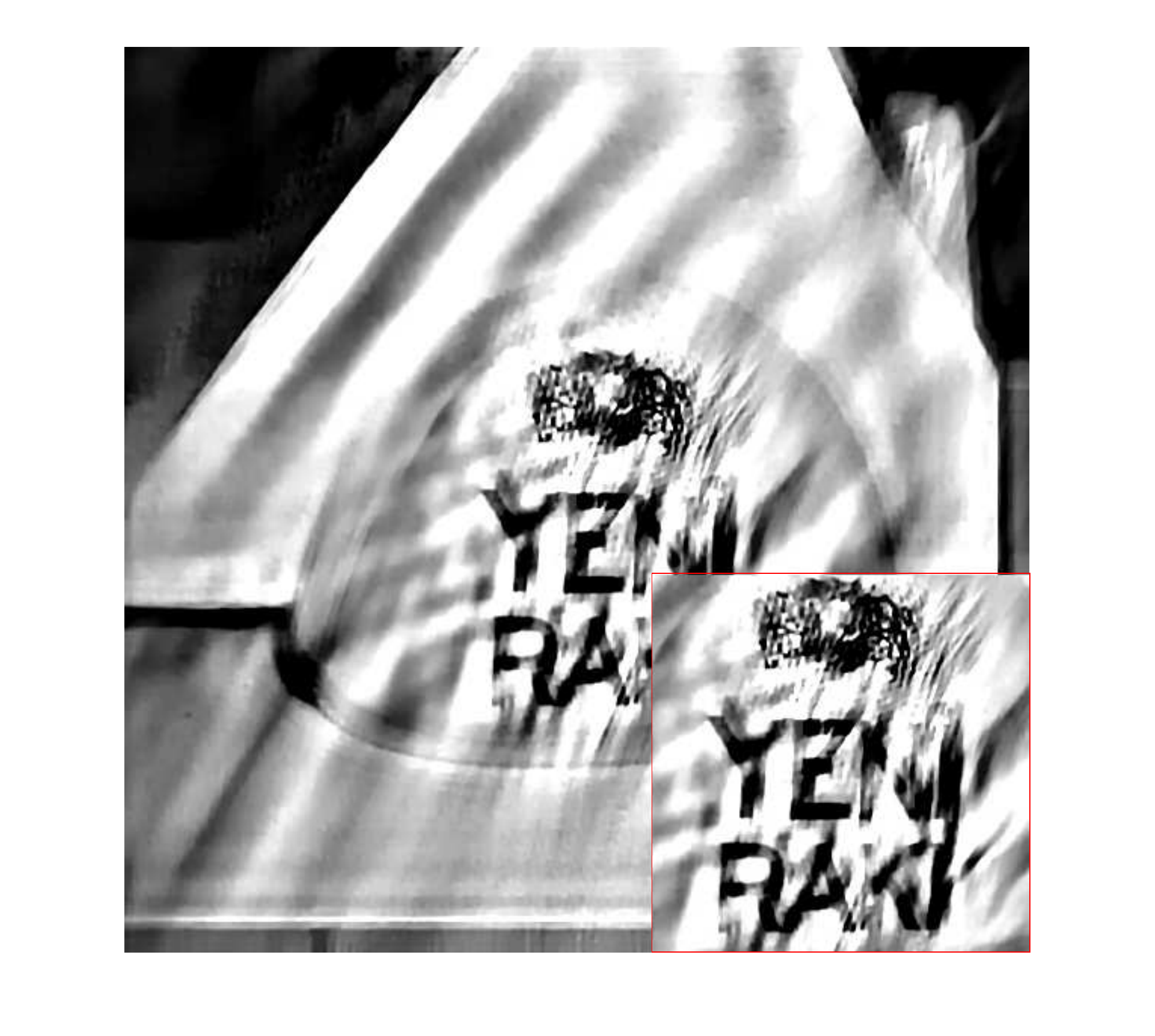}} 
	\subfloat[\scriptsize{PSNR=20.51 SSIM=0.79}]{\includegraphics[width = 1.35 in, height =1.5 in]{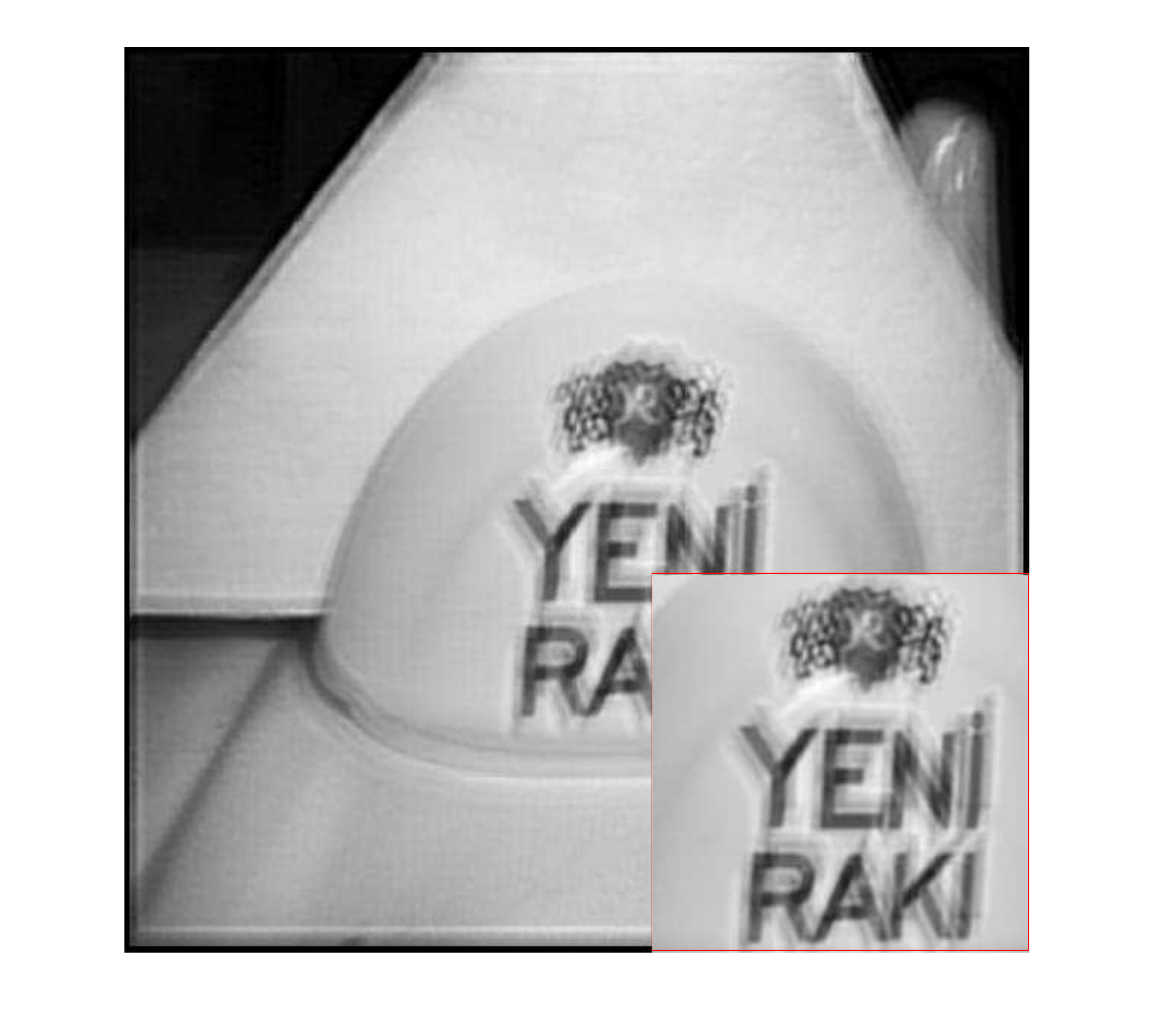}} 
	\subfloat[\scriptsize{PSNR=30.37 SSIM=0.89}]{\includegraphics[width = 1.35 in, height =1.5 in]{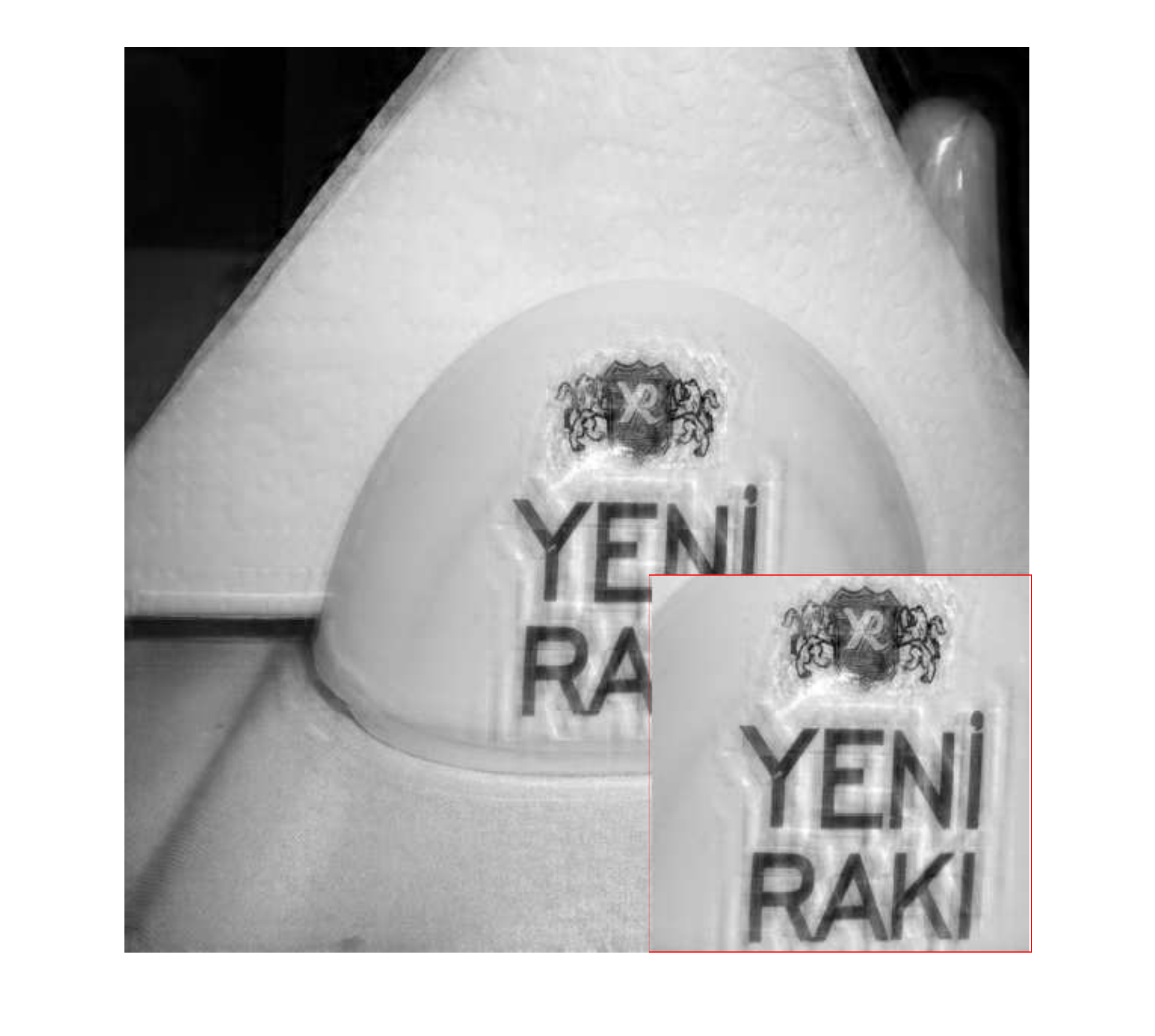}} 
	\caption{Non-uniform Blur Removal Comparison. (a) Ground truth (b) Blur kernel \cite{levin2009understanding} (c) Blurred (d) Krishnan et al. \cite{krishnan2011blind} (e) Whyte et al. \cite{whyte2014deblurring} (f) Pan et al. \cite{pan2014deblurring} (g) Xu et al. \cite{xu2013unnatural} (h) Xu et al. \cite{xu2010two} (i) Ren et al. \cite{ren2015vectorization} (j) Proposed  }
\label{nonUniform}
\end{figure*}

For visualizing the results, we show two sets of testing images. Figure \ref{Uniform} and Figure \ref{nonUniform} show the result comparison on uniform blur and non-uniform blur removal problem respectively.

\section{Conclusion}
In this paper, we proposed an optimal formulation of learning coupled autoencoders while simultaneously learning a mapping between source and target autoencoders. Our method is generalized and is applicable for any tranfer learning problem. Through experimental evaluation we showed success of our proposed method on motion blurred images. Our method is computationaly inexpensive and deblur images in seconds.\pagebreak

\bibliographystyle{IEEEbib}
\bibliography{strings,refs}

\end{document}